\documentclass{article}
\usepackage[preprint,nonatbib]{style}
\usepackage[utf8]{inputenc}
\usepackage[T1]{fontenc}
\usepackage{hyperref}
\usepackage{url}
\usepackage{booktabs}
\usepackage{amsfonts}
\usepackage{nicefrac}
\usepackage{microtype}
\usepackage{xcolor}
\usepackage[pdftex]{graphicx}
\usepackage{wrapfig}
\usepackage{multirow}
\usepackage{amsmath}
\usepackage{array}

\title{A SARS-CoV-2 Interaction Dataset and VHH Sequence Corpus for Antibody Language Models}

\author{
  Hirofumi Tsuruta$^{1,2}$,~Hiroyuki Yamazaki$^{1,3}$,~Ryota Maeda$^{1,3}$,\\
  \textbf{Ryotaro Tamura}$^{1,2}$\textbf{,~Akihiro Imura}$^{1,3}$\\
  $^1$COGNANO Inc.,~$^2$SAKURA internet Inc.,~$^3$Biorhodes, Inc.\\
  \texttt{\{tsuruta, yamazaki, maeda, ryotarotamura, akihiroimura\}@cognano.co.jp}
}

\begin{document}

\maketitle

\begin{abstract}
Antibodies are crucial proteins produced by the immune system to eliminate harmful foreign substances and have become pivotal therapeutic agents for treating human diseases.
To accelerate the discovery of antibody therapeutics, there is growing interest in constructing language models using antibody sequences.
However, the applicability of pre-trained language models for antibody discovery has not been thoroughly evaluated due to the scarcity of labeled datasets.
To overcome these limitations, we introduce AVIDa-SARS-CoV-2, a dataset featuring the antigen-variable domain of heavy chain of heavy chain antibody (VHH) interactions obtained from two alpacas immunized with severe acute respiratory syndrome coronavirus 2 (SARS-CoV-2) spike proteins.
AVIDa-SARS-CoV-2 includes binary labels indicating the binding or non-binding of diverse VHH sequences to 12 SARS-CoV-2 mutants, such as the Delta and Omicron variants.
Furthermore, we release VHHCorpus-2M, a pre-training dataset for antibody language models, containing over two million VHH sequences.
We report benchmark results for predicting SARS-CoV-2-VHH binding using VHHBERT pre-trained on VHHCorpus-2M and existing general protein and antibody-specific pre-trained language models.
These results confirm that AVIDa-SARS-CoV-2 provides valuable benchmarks for evaluating the representation capabilities of antibody language models for binding prediction, thereby facilitating the development of AI-driven antibody discovery.
The datasets are available at \url{https://datasets.cognanous.com}.
\end{abstract}

\section{Introduction}

Antibodies are vital proteins produced by the immune system to remove harmful foreign substances called antigens.
Antibody-based therapeutics, which can bind to target antigens with high affinity and specificity, have become a major class of therapeutic agents and are currently used to treat a wide range of diseases~\cite{al2022therapeutic,alejandra2023production}.
Among their successes, the rapid development and subsequent approval of antibodies against severe acute respiratory syndrome coronavirus 2 (SARS-CoV-2) epitomize the impactful response of this therapeutic class in addressing urgent global health challenges~\cite{zhou2020perspectives,li2023therapeutic}.
However, the development of therapeutic antibodies remains a time-consuming and costly endeavor due to the complexity and difficulty of artificially manipulating the vast search space of antibody sequences~\cite{kandari2023antibody}.
Therefore, computational approaches for accelerating antibody discovery have become increasingly popular in recent years~\cite{wilman2022machine,kim2023computational,bai2023accelerating}.

Recent advances in language models offer new possibilities for understanding the information contained in antibody sequences because an antibody sequence can be represented as a string of letters representing a type of amino acid.
With the construction of the observed antibody space (OAS) database~\cite{kovaltsuk2018observed,olsen2022observed} that currently contains over two billion antibody sequences, a sufficient number of antibody sequences is now available to train antibody-specific language models~\cite{ruffolo2021deciphering,leem2022deciphering,olsen2022ablang,wang2022pre,porebski2024rapid,barton2024enhancing,kenlay2024large}.
Olsen \textit{et al.}~\cite{olsen2022ablang} presented AbLang, an antibody language model pre-trained on either the heavy or light chain antibody sequences in the OAS database.
They demonstrated that AbLang can be used to accurately restore the missing residues in antibody sequences.
Wang \textit{et al.}~\cite{wang2022pre} proposed EATLM, a pre-trained antibody language model that incorporates evolutionary information as the pre-training objectives.
They also provided an antibody understanding evaluation (ATUE) benchmark consisting of four tasks to evaluate the performance of pre-trained language models in antibody-related tasks.

Despite these promising developments, the applicability of pre-trained language models for antibody discovery has not been adequately evaluated due to the lack of labeled datasets.
ATUE includes an antibody discovery task, a binary sequence classification that distinguishes antibodies that bind to SARS-CoV-2.
The training dataset for this task used antibody sequences from SARS-CoV-2 patients and healthy persons from the OAS database.
Although very few antibodies from SARS-CoV-2 patients are directly responsible for virus binding, these noisy and potentially unreliable individual-level disease labels were used to train a sequence-level classifier.
Thus, a dataset with labels indicating whether the antibody binds to a specific antigen at the antibody sequence level would be extremely useful for a more accurate evaluation of model performance for antibody discovery.

In this study, we introduce AVIDa-SARS-CoV-2, a dataset featuring the antigen-variable domain of heavy chain of heavy chain antibody (VHH) interactions produced by two alpacas immunized with SARS-CoV-2 spike proteins.
VHHs, found in camelids such as alpacas and llamas, are promising therapeutic agents because of their small size, high stability, and high antigen-binding affinity~\cite{jovvcevska2020therapeutic,jin2023nanobodies}.
AVIDa-SARS-CoV-2 was generated using our previously established method for generating interaction datasets with reliable labels~\cite{tsuruta2023avida}.
AVIDa-SARS-CoV-2 contains binary labels that indicate whether each of the diverse VHH sequences binds or does not bind to 12 SARS-CoV-2 mutants, such as the Delta and Omicron variants.
Notably, label reliability was verified by experimental evidence that VHHs extracted from AVIDa-SARS-CoV-2 bound to SARS-CoV-2 spike variants~\cite{maeda2022panel}.

Furthermore, we introduce VHHCorpus-2M, a pre-training dataset for antibody language models containing over two million VHH sequences, and VHHBERT, an antibody language model pre-trained on VHHCorpus-2M.
To avoid sequencing errors and increase sequence reliability, we removed singletons from the VHH sequences identified by next-generation sequencing (NGS), that is, only sequences observed more than once were used in the corpus.
Although VHHCorpus-2M contains fewer sequences than OAS, it is distinctive in that it consists entirely of full-length VHH sequences that act as the smallest functional units for binding to each target antigen.

The main contributions of this paper are summarized as follows.

\begin{itemize}
  \item We release AVIDa-SARS-CoV-2, a labeled SARS-CoV-2-VHH interaction dataset with amino acid sequences, and VHHCorpus-2M, which contains over two million unlabeled VHH sequences.
  These datasets can be used for the evaluation and pre-training of antibody-specific language models.
  \item AVIDa-SARS-CoV-2 contains information on the interactions of diverse VHHs produced by two alpacas with 12 SARS-CoV-2 mutants, providing researchers with valuable insights into the effects of antigen mutations on antibody binding and individual differences in antigen-specific VHHs.
  \item We release VHHBERT, a VHH-specific language model pre-trained using VHHCorpus-2M.
  VHHBERT will serve as a baseline for subsequent VHH-specific language models.
  \item We report benchmark results for the prediction of SARS-CoV-2-VHH interactions using VHHBERT and existing general protein and antibody-specific pre-trained language models.
  These results confirm that AVIDa-SARS-CoV-2 provides valuable benchmarks for assessing the representation capabilities of antibody language models for binding prediction.
\end{itemize}

\section{Related Work}

\begin{table}
  \caption{Characteristics of pre-trained antibody language models.
  ``M'' stands for million, and ``B'' stands for billion.}
  \centering
  \renewcommand{\arraystretch}{1.1}
  \label{tab:palm_list}
  \resizebox{\textwidth}{!}{
  \begin{tabular}{cccccccc}
    \toprule
    & \multicolumn{3}{c}{Pre-training} && \multicolumn{3}{c}{Evaluation} \\ \cline{2-4} \cline{6-8}
    Model & Dataset & \#Samples & Chain Type && Dataset & \#Samples & Task \\
    \toprule
    AntiBERTy~\cite{ruffolo2021deciphering} & OAS & 588M & Heavy, light && HIV-1 donor repertoires~\cite{zhou2013multidonor,zhou2015structural} & 232,593 & Evolutionary analysis \\ \hline
    AntiBERTa~\cite{leem2022deciphering} & OAS & 72M & Heavy, light && SAbDab~\cite{dunbar2014sabdab} & 900 & Paratope prediction \\ \hline
    AbLang-H~\cite{olsen2022ablang} & OAS & 14M & Heavy && OAS & 2,000 & Sequence restoration \\ \hline
    AbLang-L~\cite{olsen2022ablang} & OAS & 0.24M & Light && OAS & 4,200 & Sequence restoration \\ \hline
    EATLM~\cite{wang2022pre} & OAS & 20M & Heavy, light && 
      \begin{tabular}{c}  
        Mason \textit{et al.}'s dataset~\cite{mason2021optimization}\\SAbDab~\cite{dunbar2014sabdab}\\Mroczek \textit{et al.}'s dataset~\cite{mroczek2014differences}\\OAS, CoV-AbDab~\cite{raybould2021cov}
      \end{tabular} & 
      \begin{tabular}{c}
        21,612\\1,662\\88,094\\22,000
      \end{tabular} &
      \begin{tabular}{c}
      Binding prediction\\Paratope prediction\\B cell classification\\Antibody discovery
      \end{tabular} \\ \hline
    BERT-DS~\cite{porebski2024rapid} & OAS & 20M & Heavy && HER2affmat & 234,088 & Binding prediction \\ \hline
    AntiBERTa2~\cite{barton2024enhancing} & 
      \begin{tabular}{c}
        OAS,\\proprietary dataset
      \end{tabular} & 
      824M & Heavy, light && Mason \textit{et al.}'s dataset~\cite{mason2021optimization} & 22,779 & Binding prediction \\ \hline
    IgBert~\cite{kenlay2024large} & OAS & 2B & Heavy, light && 
    \begin{tabular}{c}  
      OAS\\FLAb~\cite{chungyoun2024flab}\\OAS
    \end{tabular} & 
    \begin{tabular}{c}
      20,000\\6,745\\1,000
    \end{tabular} &
    \begin{tabular}{c}
    Sequence restoration\\Binding affinity prediction\\Perplexity
    \end{tabular} \\ \hline
    \bf{VHHBERT} & \bf{VHHCorpus-2M} & \bf{2M} & \bf{Heavy} && \bf{AVIDa-SARS-CoV-2} & \bf{77,003} & \bf{Binding prediction} \\
    \bottomrule
  \end{tabular}
  }
\end{table}

In this section, we put our work in the context of existing pre-trained antibody language models and their datasets used for pre-training and evaluation.
Currently, there is growing interest in constructing language models using protein sequences~\cite{rives2021biological,rao2021msa,elnaggar2022prottrans,lin2023evolutionary}.
Inspired by these successes and the fact that the evolutionary process of antibodies is significantly different from that of proteins, several studies have attempted to train language models specific to antibody sequences.
The representative existing studies are summarized in Table~\ref{tab:palm_list}.

\paragraph{Pre-trained Antibody Language Models.}

Ruffolo \textit{et al.}~\cite{ruffolo2021deciphering} proposed AntiBERTy, the first antibody-specific language model to understand affinity maturation within immune repertoires.
They found that AntiBERTy can cluster antibodies into trajectories resembling affinity maturation.
Leem \textit{et al.}~\cite{leem2022deciphering} presented a pre-trained antibody language model called AntiBERTa and fine-tuned AntiBERTa to predict the binding site of an antibody, called a paratope, from an antibody sequence.
Olsen \textit{et al.}~\cite{olsen2022ablang} pre-trained two language models: one trained only on heavy chains of antibodies (AbLang-H) and one trained only on light chains of antibodies (AbLang-L).
These models outperformed ESM-1b~\cite{rives2021biological}, a general protein language model, in restoring the missing residues in antibody sequences.
Wang \textit{et al.}~\cite{wang2022pre} incorporated two original pre-training objectives into their proposed antibody language model, EATLM, to explore the benefits of incorporating specific biological mechanisms into pre-training.
They also provided a useful benchmark, ATUE, that consists of four antibody-related tasks.
Porebski \textit{et al.}~\cite{porebski2024rapid} pre-trained BERT-DS using 20 million heavy-chain human antibody sequences and fine-tuned it for binding prediction.
Barton \textit{et al.}~\cite{barton2024enhancing} developed AntiBERTa2, an antibody language model pre-trained using 824 million antibody sequences including paired heavy and light chain sequences, and proposed a multimodal contrastive learning that amalgamates the representations of antibody sequences and structures.
Kenlay \textit{et al.}~\cite{kenlay2024large} presented IgBert, which was initialized with the pre-trained protein language model ProtBERT~\cite{elnaggar2022prottrans} and trained using more than two billion unpaired antibody sequences (i.e., heavy chain or light chain only) and two million paired antibody sequences.
Various other antibody-specific language models have been developed for antibody humanization~\cite{prihoda2022biophi}, sequence generation~\cite{shuai2021generative,nijkamp2023progen2,melnyk2023reprogramming}, identification of evolutionarily plausible mutations~\cite{hie2024efficient}, and classification of antigen-specific antibodies~\cite{burbach2024improving}.

\paragraph{Pre-training Datasets.}

The pre-training datasets for antibody language models are large collections of unlabeled antibody sequences.
Conventional antibodies in humans and mice comprise two pairs of heavy and light chains, meaning that one pair of chains serves as the functional unit for binding to the target antigen.
Currently, OAS~\cite{kovaltsuk2018observed,olsen2022observed} contains over two billion unpaired antibody sequences, more than 90\% of which are of human origin.
Existing antibody language models are pre-trained primarily using unpaired antibody sequences in the OAS database.
AntiBERTa2 and IgBert were trained with paired antibody sequences, but with a very small proportion compared with unpaired sequences.

VHHs are variable regions of heavy-chain antibodies found in camelids.
Because VHH acts as a single functional unit, its sequence contains all the information necessary for antibody functions against antigens.
The OAS database currently contains approximately 1.6 million unique VHH sequences collected from one study~\cite{li2016comparative} and derived from three unimmunized Bactrian camels.
The Integrated Nanobody Database for Immunoinformatics (INDI) database~\cite{deszynski2022indi} currently contains more than 11 million unique VHH sequences collected mainly from the Sequence Read Archive (SRA)~\cite{leinonen2010sequence} and derived from dromedaries, Bactrian camels, llamas, and alpacas.
VHHCorpus-2M contains more than two million unique VHH sequences generated in original experiments using five alpacas.

\paragraph{Evaluation Datasets.}

The evaluation datasets are a set of labeled antibody sequences used to assess model performance for a specific task.
AntiBERTa and EATLM were fine-tuned using the structural antibody database (SAbDab)~\cite{dunbar2014sabdab} to evaluate their performance in predicting paratopes.
Paratope prediction is important for the efficient discovery of antibody candidates that bind to an antigen of interest; however, the size of labeled datasets is limited.
IgBert was evaluated in a binding affinity prediction task using each of three datasets with small data samples of 422~\cite{shanehsazzadeh2023unlocking}, 2048~\cite{warszawski2019optimizing}, and 4275~\cite{koenig2017mutational} from the fitness landscape for antibodies (FLAb)~\cite{chungyoun2024flab}.
BERT-DS was evaluated in a binding prediction task involving a three-category classification using a deep-screening dataset called HER2affmat.
Although HER2affmat is a useful dataset with a large number of samples, it does not contain full-length antibody sequences.
A binding prediction task using Mason \textit{et al.}'s dataset~\cite{mason2021optimization} in ATUE~\cite{wang2022pre} was done that involved binary classification to determine whether the complementarity-determining region (CDR) of an antibody can bind to human epidermal growth factor receptor 2 (HER2).
AntiBERTa2 was also evaluated for its performance in binding prediction using the same dataset.
All antibody sequences in this dataset were derived from a single germline sequence, indicating that the diversity of antibody sequences was strongly limited.
Thus, in ATUE, this task is considered to be less relevant to antibody-specific evolution.

The antibody discovery task in ATUE is a binary sequence classification that distinguishes antibodies that bind to SARS-CoV-2.
This task has two notable limitations in terms of accurate model evaluation for antibody discovery.
First, the dataset for training the sequence classifier uses antibody sequences with noisy individual-level labels from SARS-CoV-2 patients and healthy persons, even though very few antibodies from SARS-CoV-2 patients are responsible for virus binding.
Second, this task assumes that if the third CDR of the heavy chain (CDR-H3) of the binder sequence predicted by the model is 90\% or more identical to the CDR-H3 of the true binding sequences in CoV-AbDab~\cite{raybould2021cov}, they have a similar binding performance.
However, not only the sequence of CDRs but also the appropriate three-dimensional structure and interactions between variable regions are important for antigen-antibody activity~\cite{nakanishi2008critical}.
AVIDa-SARS-CoV-2 has sequence-level labels for binding and non-binding to SARS-CoV-2 mutants for each full-length VHH sequence.

\section{AVIDa-SARS-CoV-2: Antigen-VHH Interaction Dataset Produced from Alpaca Immunized with SARS-CoV-2 Spike Proteins}
\label{sec:proposal_avida}

AVIDa-SARS-CoV-2 is an antigen-VHH interaction dataset with 77,003 data samples, comprising 22,002 binding pairs and 55,001 non-binding pairs.
The dataset was released under a CC BY-NC 4.0 license and is available at \url{https://avida-sars-cov-2.cognanous.com}.

\subsection{Dataset Generation}
\label{sec:avida_data_generation}

\begin{figure}
  \centering
  \includegraphics[width=\textwidth]{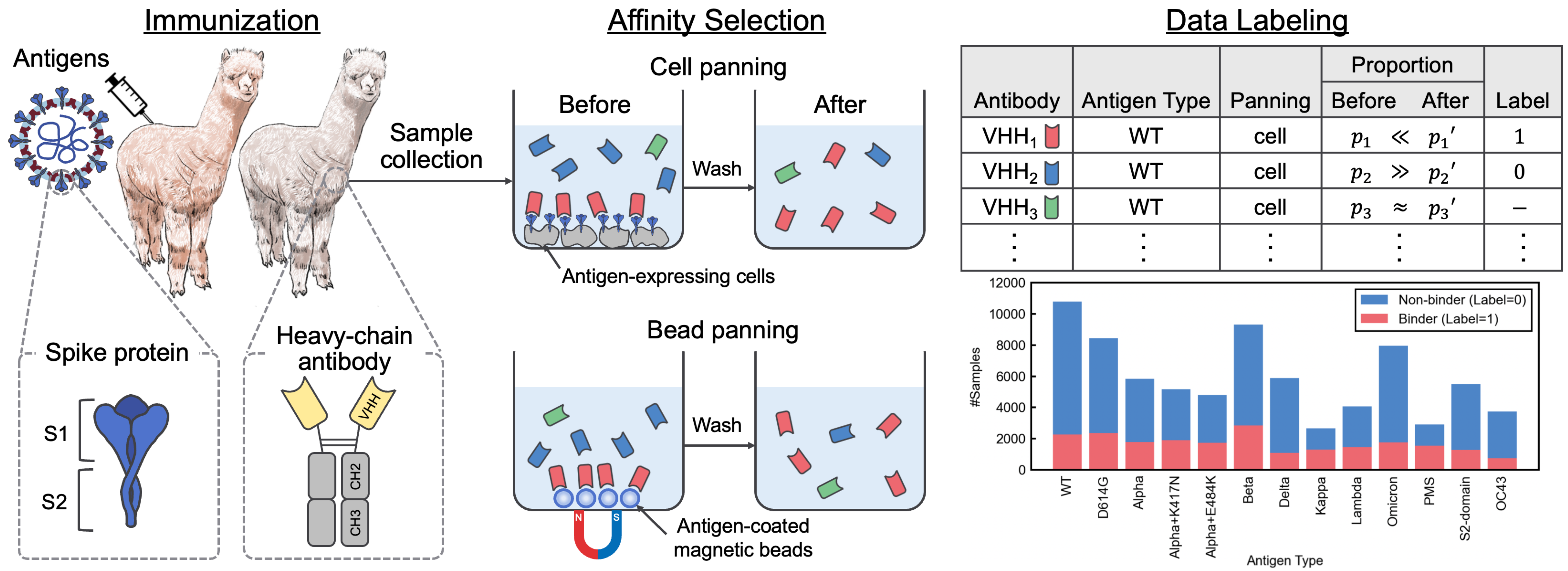}
  \caption{Overview of data generation process for AVIDa-SARS-CoV-2.}
  \label{fig:data_generation_overview}
\end{figure}

AVIDa-SARS-CoV-2 was generated using a method established in our previous study~\cite{tsuruta2023avida}.
This section introduces the overall workflow and key concepts underlying our data generation, as shown in Figure~\ref{fig:data_generation_overview}.
Appendix \ref{sec:appendix_dataset_generation} provides the detailed step-by-step procedures for dataset generation.

\paragraph{Immunization}

\begin{table}
  \caption{Summary of antigen types.
  Appendix~\ref{sec:appendix_dataset_generation} gives more details on each antigen.
  }
  \centering
  \label{tab:antigen_list}
  \resizebox{110mm}{!}{
  \begin{tabular}{ccl}
    \toprule
    Antigen Type & Panning & Description \\
    \midrule
    WT & cell & Wild-type (\textbf{WT}) SARS-CoV-2 identified in Wuhan \\
    D614G & cell & Mutant with \textbf{D614G} mutation \\
    Alpha & cell, bead & Mutant with representative mutations of \textbf{Alpha} variant \\
    Alpha+K417N & cell & Mutant of antigen type ``Alpha'' with \textbf{K417N} mutation \\
    Alpha+E484K & cell & Mutant of antigen type ``Alpha'' with \textbf{E484K} mutation \\
    Beta & cell, bead & Mutant with representative mutations of \textbf{Beta} variant \\
    Delta & cell, bead & Mutant with representative mutations of \textbf{Delta} variant \\
    Kappa & bead & Mutant with representative mutations of \textbf{Kappa} variant \\
    Lambda & bead & Mutant with representative mutations of \textbf{Lambda} variant \\
    Omicron & cell, bead & Mutant with representative mutations of \textbf{Omicron} (BA.1) variant \\
    PMS & bead & Polymutant spike (\textbf{PMS}) protein~\cite{schmidt2021high} \\
    S2-domain & bead & \textbf{S2-domain} of the WT \\
    OC43 & bead & Human coronavirus \textbf{OC43} (HCoV-OC43) \\
    \bottomrule
  \end{tabular}
  }
\end{table}

We used the immune system of live alpacas to obtain diverse VHHs that bind to SARS-CoV-2.
First, we immunized two alpacas (hereafter referred to as Alpaca P and Alpaca C) that were maternal half-siblings with the 13 types of antigens listed in Table~\ref{tab:antigen_list}.
The spike protein of SARS-CoV-2, which protrudes from the virus surface, is a crucial structural component that facilitates its entry into host cells by binding to receptors on cells.
Owing to its crucial role in the infection process, the spike protein is the primary target for antibodies.
As the virus evolves over time, mutations in the spike protein that escape the immune response are enriched, and the effectiveness of antibodies to neutralize the virus is reduced.
To investigate the effects of mutations in the spike protein, we generated mutants by selecting representative mutations that are effective for immune escape among the mutations observed to date.

\paragraph{Affinity Selection}

After immunization, an alpaca's body harbors a small amount of SARS-CoV-2-specific VHHs produced by the immune response and a large amount of VHHs unrelated to SARS-CoV-2.
To distinguish between them, we performed affinity selection by biopanning using the spike proteins listed in Table~\ref{tab:antigen_list} as target molecules.
We performed either bead panning, cell panning, or both for each target molecule.
For bead panning, the ectodomain of the spike protein was produced by cells, purified, and then combined with beads as bait for panning.
For cell panning, the bait was a whole cell overexpressing the full-length spike proteins on the cell membrane with the ectodomain protruding out.
Through this process, target-specific VHHs become enriched, while non-specific VHHs are gradually diluted out, ultimately yielding a concentrated sample of target-specific VHHs.

\paragraph{Data Labeling}

We counted the number of occurrences of each unique VHH sequence in the samples before and after affinity selection by NGS, which reflected the proportion of each VHH in the samples.
We then compared the proportions of each VHH before and after affinity selection and labeled VHHs whose proportions significantly increased as ``binder'' and VHHs whose proportions significantly decreased as ``non-binder'' on the basis of statistical tests.
In addition, VHHs whose proportions did not change significantly, corresponding to about 97\% of the total, were excluded from the dataset to improve label reliability.
We previously verified the reliability of this labeling method by confirming the binding ability of 20 labeled VHHs in AVIDa-hIL6~\cite{tsuruta2023avida} using immunofluorescence staining analysis and biolayer interferometry analysis.
Furthermore, label reliability was supported by experimental evidence that nine VHHs extracted from AVIDa-SARS-CoV-2 bound to SARS-CoV-2 spike variants, including Omicron~\cite{maeda2022panel}.

\subsection{Dataset Analysis}
\label{sec:avida_dataset_analysis}

\paragraph{Binding Sensitivity to Sequence Variation}

\begin{figure}
  \centering
  \includegraphics[width=\textwidth]{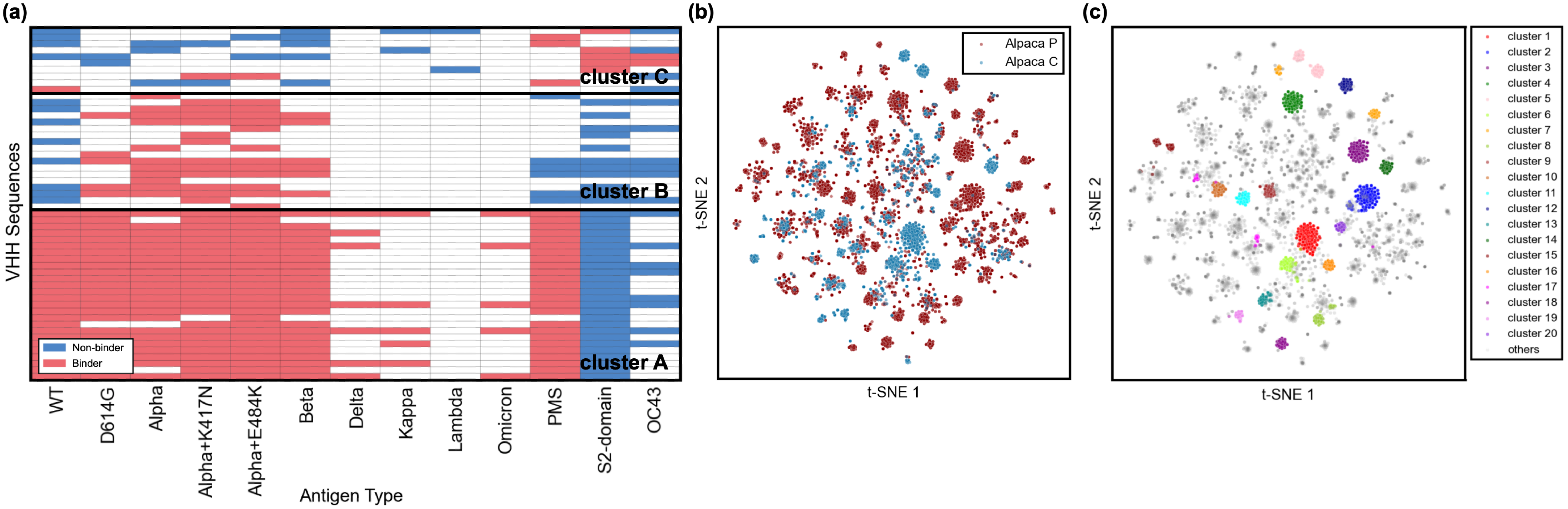}
  \caption{(a) Label visualization for each pair between 54 VHHs in three clusters and antigens.
  Each cell represents unique VHH-antigen pair.
  White cells are unlabeled pairs that cannot be identified as ``binder'' or ``non-binder'' and are not included in AVIDa-SARS-CoV-2.
  (b)(c) Two-dimensional representation of binder sequences colored by individuals and clusters.
  Appendix~\ref{sec:appendix_dataset_analysis} provides enlarged versions of (b) and (c).}
  \label{fig:avida_dataset_analysis}
\end{figure}

AVIDa-SARS-CoV-2 contains information regarding whether the same VHH sequence binds to each antigen type.
The number of unique VHH sequences in AVIDa-SARS-CoV-2 is 36,100 including 14,078 sequences that bind to at least one antigen type.
Notably, 427 VHH sequences were labeled as ``binder'' to specific antigen types but ``non-binder'' to others.
In the case of infectious diseases and malignancies, the target antigen can mutate to escape the immune system or develop tolerance to treatment.
If the binding site of an antigen, called an epitope, is mutated, the corresponding antibody will no longer bind to it.
Conversely, if an antibody that loses binding activity due to an antigen mutation is identified, the location of the mutation can be assumed to be close to the epitope.
Therefore, we further examined the binding activity of these 427 VHH sequences.

First, we clustered 427 VHH sequences using MMseqs2~\cite{steinegger2017mmseqs2} with 90\% sequence identity, resulting in 38 clusters of size two or more.
We then extracted 54 VHH sequences from the three clusters in descending order of cluster size and visualized whether each VHH sequence bound to each antigen type, as shown in Figure~\ref{fig:avida_dataset_analysis}(a).
Focusing on the vertical direction in cluster C, some sequences with over 90\% sequence identity can exhibit varying binding abilities against the same antigen.
Focusing on the horizontal direction, it is clear that the same VHH has different binding abilities for different antigen types.
For example, in cluster B, antigen types from WT to Beta, which differ by only a few amino acids, can alter VHH binding, suggesting that these mutations enhance or inhibit binding.
Interestingly, all VHHs in cluster A bind to WT but not to S2-domain, indicating that these VHHs bind to the S1 region of the spike protein.
We can also recognize that most of these VHHs cannot be identified as binders for variants such as Delta, Kappa, Lambda, and Omicron, which exactly reflects the immune escape phenomenon in the real world.
Therefore, AVIDa-SARS-CoV-2 contains sensitive information in which small amino acid sequence variations of an antibody and antigen can change between binding or non-binding, which should be strongly associated with their binding sites.

\paragraph{Individual Differences in Antigen-specific Antibody Production}

We compared the differences in SARS-CoV-2-specific VHHs produced by the immune systems of the two alpacas.
The number of unique VHH binders for Alpaca P and Alpaca C were 10,487 and 3,651, respectively, of which 60 VHHs were observed in both individuals.
We encoded VHH sequences using Kidera factors~\cite{kidera1985statistical}, which represent the physicochemical properties of amino acids in a 10-dimensional vector, and then converted them into two-dimensional (2D) vectors using t-SNE~\cite{van2008visualizing}.
Figure~\ref{fig:avida_dataset_analysis}(b) shows a 2D representation of the VHH binders.
The data points derived from each individual partially overlapped but predominantly aggregated in distinct regions.
Figure~\ref{fig:avida_dataset_analysis}(c) shows a 2D representation of the VHH binders clustered using MMseqs2 with 95\% sequence identity and colored into 20 clusters in descending order of cluster size.
This result indicates that the aggregations in 2D space reflect the VHH clusters formed on the basis of sequence identity.
For example, cluster 1 (colored red) is composed of VHHs produced from Alpaca C, whereas cluster 2 (colored blue) is composed of VHHs produced from Alpaca P.
These results demonstrate that using multiple individuals in dataset generation contributes to enhancing the diversity of antigen-specific VHH sequences.

\paragraph{Differences with AVIDa-hIL6}

Building on the findings from the above analysis, we elucidate the differences between AVIDa-SARS-CoV-2 and the previously released AVIDa-hIL6~\cite{tsuruta2023avida}, beyond the target antigens used for immunization.
AVIDa-hIL6 used human interleukin-6 (IL-6) mutants produced by artificial point mutations, whereas AVIDa-SARS-CoV-2 used SARS-CoV-2 spike proteins with natural mutations that are more important for antigen-antibody interactions.
This allowed AVIDa-SARS-CoV-2 to contain labels that reflect the immune escape phenomenon in the real world, as shown in Figure~\ref{fig:avida_dataset_analysis}(a).
Moreover, AVIDa-hIL6 collected VHHs from one alpaca, whereas AVIDa-SARS-CoV-2 collected them from two alpacas.
This increased the diversity of antigen-specific antibodies and provided valuable insights into the sequence differences of antigen-specific antibodies between individuals, as shown in Figure~\ref{fig:avida_dataset_analysis}(b).

\section{VHHCorpus-2M: VHH Sequence Corpus Produced from Alpaca}
\label{sec:proposal_corpus}

VHHCorpus-2M is a corpus containing 2,040,988 unique VHH sequences.
The corpus was released under a CC BY-NC 4.0 license and is available at \url{https://vhh-corpus.cognanous.com}.

\subsection{Dataset Collection}
\label{sec:corpus_data_collection}

VHHCorpus-2M is a collection of unique VHH sequences from several datasets generated by the process described in Section~\ref{sec:avida_data_generation} using target antigens other than the SARS-CoV-2 spike protein, such as the human immunodeficiency virus type 1 (HIV-1) envelope protein, human IL-6, HER2, human histone, transmembrane glycoprotein mucin 1 (MUC1), and gram-negative bacteria.
We collected VHH sequences from datasets produced by five alpacas, different from those used in the generation of AVIDa-SARS-CoV-2, to avoid potential data leakage and increase the diversity of VHH sequences.
Note that the source datasets include publicly available AVIDa-hIL6~\cite{tsuruta2023avida} in addition to multiple datasets that have not been published as labeled binding datasets.
Importantly, we used only VHH sequences that were identified multiple times by NGS in our corpus to avoid sequencing errors and increase sequence reliability.

\subsection{Dataset Analysis}
\label{sec:corpus_dataset_analysis}

\begin{wrapfigure}[13]{r}{0.35\textwidth}
  \centering
  \vspace{-5mm}
  \includegraphics[width=0.35\textwidth]{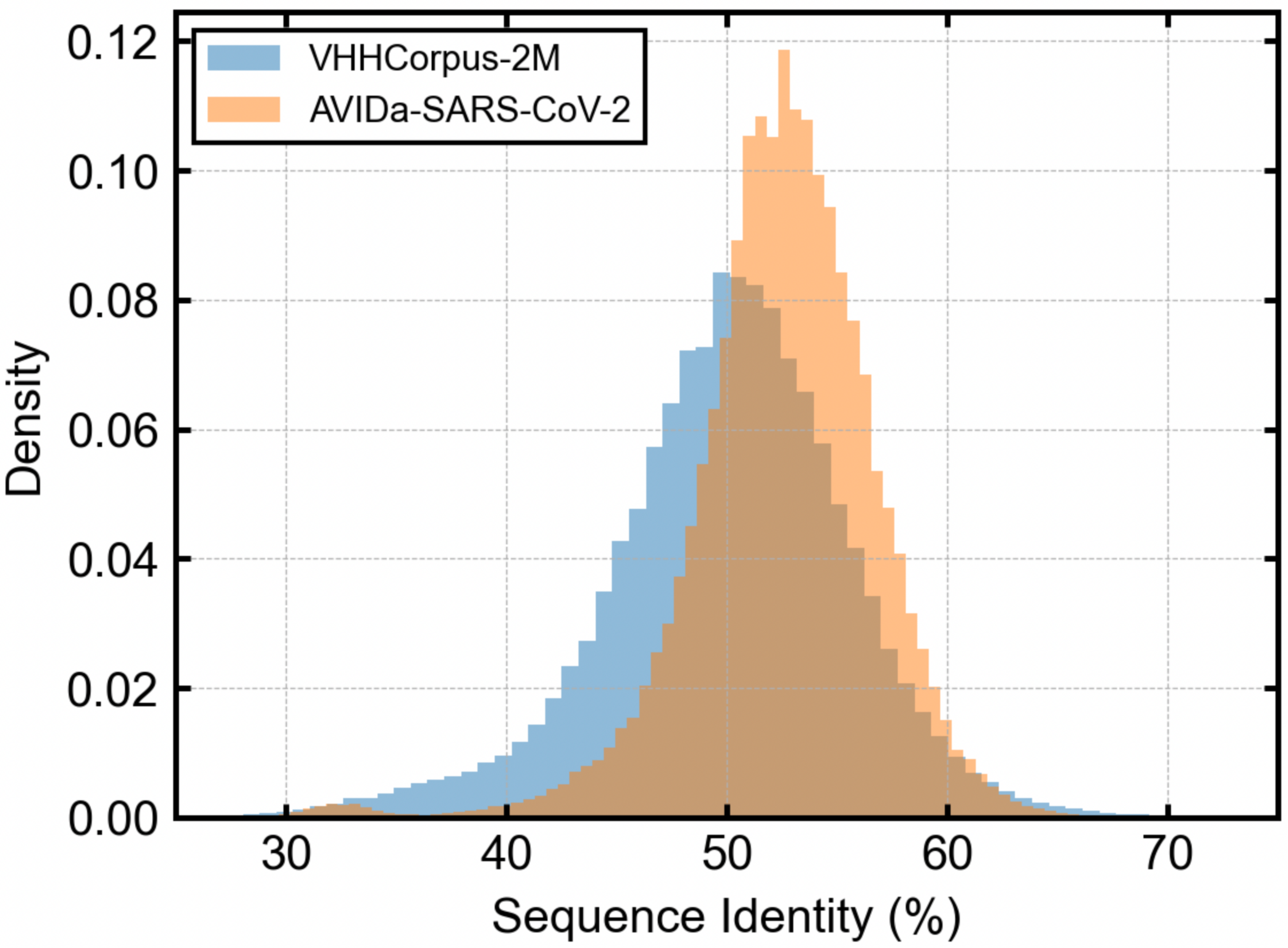}
  \caption{Distribution of pairwise identities of VHH sequences.}
  \label{fig:corpus_dataset_analysis}
\end{wrapfigure}

The size and diversity of pre-training datasets play an important role in improving the performance of a language model in downstream tasks~\cite{radford2019language,liu2019roberta}.
VHHCorpus-2M comprises 2,040,988 unique sequences, which is more than 50 times the number of unique sequences in AVIDa-SARS-CoV-2.
To examine the degree of sequence diversity in the datasets, we calculated the pairwise sequence identities within each dataset.
First, to mitigate the computational complexity, we used MMseqs2 to cluster the VHH sequences with 70\% sequence identity within each dataset, resulting in 8,270 and 777 clusters for VHHCorpus-2M and AVIDa-SARS-CoV-2, respectively.
We then extracted representative sequences from all clusters and calculated all pairwise sequence identities among these representatives for each dataset.
Figure~\ref{fig:corpus_dataset_analysis} presents the distribution of pairwise sequence identities for each dataset.
The distribution of VHHCorpus-2M has a broader peak in regions with a lower sequence identity compared with AVIDa-SARS-CoV-2, indicating a higher sequence diversity.
This could be attributed to the fact that VHHCorpus-2M originated from five alpacas, which is more than AVIDa-SARS-CoV-2.
Additionally, VHHCorpus-2M includes unlabeled VHH sequences that cannot be labeled as ``binder'' or ``non-binder'' for specific antigens, further contributing to its diversity.

\section{Benchmarks}

\subsection{Benchmark Task}
\label{sec:benchmark_task}

\begin{wraptable}[6]{r}{0.35\textwidth}
  \vspace{-8mm}
  \caption{Numbers of samples in training and test sets.}
  \centering
  \label{tab:train_test_split}
  \resizebox{50mm}{!}{
  \begin{tabular}{cccc}
    \toprule
     & \multicolumn{3}{c}{\#Samples} \\ \cline{2-4}
    Dataset & Binder & Non-binder & Total \\
    \midrule
    Training & 15,400 & 34,285 & 49,685 \\ \hline
    Test & 6,602 & 20,716 & 27,318 \\
    \bottomrule
  \end{tabular}
  }
\end{wraptable}

To evaluate the performance of various pre-trained language models for antibody discovery, we defined a binary classification task to predict the binding or non-binding of unknown antibodies to 13 antigens using AVIDa-SARS-CoV-2.
By leveraging the binding information of diverse VHHs produced from the two alpacas, we used data samples obtained from Alpaca P as the training set and data samples obtained from Alpaca C as the test set.
Table~\ref{tab:train_test_split} lists the number of samples in each set.
As shown in Figure~\ref{fig:avida_dataset_analysis}(b) and (c), the VHH binders derived from different alpacas formed distinct clusters.
Therefore, this experimental scenario assumes that we want to explore additional effective antibodies beyond those already observed to bind to a known antigen.
This scenario holds significant importance in the development of therapeutic antibodies, given that antibodies with different sequences can bind to different binding sites of antigens, called epitopes.
Depending on their binding sites, antibodies may have specific biologically important functions, such as neutralization, inhibition, or activation, and can be extremely useful in drugs.

\subsection{Experimental Settings}
\label{sec:experimental_settings}

\paragraph{Baseline Models}
\label{sec:baseline_models}

To fully evaluate the representation capabilities of the language models pre-trained on various training sequence data, we selected the following nine baseline models.
(1) \textbf{ProtBert}~\cite{elnaggar2022prottrans} is a BERT-based~\cite{devlin2018bert} model pre-trained on 216 million protein sequences in UniRef~\cite{suzek2015uniref}.
(2)(3) \textbf{ESM-2}~\cite{lin2023evolutionary} is a model pre-trained on 65 million unique protein sequences in UniRef~\cite{suzek2015uniref}.
We used ESM-2 with 150 and 650 million parameters (hereafter referred to as ESM-2 150M and ESM-2 650M).
We adopted ProtBert and ESM-2 to confirm whether pre-training with antibodies, a subset of proteins, is effective for predicting VHH binding.
(4) \textbf{AbLang-H}~\cite{olsen2022ablang} is a RoBERTa-based~\cite{liu2019roberta} model pre-trained on 14 million heavy chains of antibodies in the OAS database.
(5) \textbf{AntiBERTa2}~\cite{barton2024enhancing} is a RoFormer-based~\cite{su2024roformer} model pre-trained using 824 million antibody sequences including paired antibody sequences in the OAS and proprietary database.
(6) \textbf{AntiBERTa2-CSSP}~\cite{barton2024enhancing} is a multimodal version of AntiBERTa2 that is further trained on human antibody structures using contrastive sequence-structure pre-training (CSSP).
(7) \textbf{IgBert}~\cite{kenlay2024large} is a model initialized with weights of ProtBert and trained using more than two billion unpaired sequences of light and heavy chains and two million paired sequences in the OAS database.
(8) \textbf{VHHBERT} is a RoBERTa-based model pre-trained on two million VHH sequences in VHHCorpus-2M.
We used the same model parameters as RoBERTa\textsubscript{BASE}, except that it used positional embeddings with a length of 185 to cover the maximum sequence length of 179 in VHHCorpus-2M.
(9) \textbf{VHHBERT w/o PT} is a VHHBERT initialized with random weights without pre-training.
We adopted this model to confirm the effectiveness of pre-training.

\paragraph{Pre-training}

As a pre-training corpus for VHHBERT, VHHCorpus-2M was randomly divided into 2,000,000 training sets and 40,988 validation sets.
The VHH sequences were tokenized by mapping each of the 20 amino acids to a different token ID and adding special tokens at the beginning and end of the sequence.
We used masked language modeling as the pre-training objective.
During pre-training, 15\% of the residues from each VHH sequence were randomly selected, and of these, 80\% were masked, 10\% were randomly changed to another residue, and 10\% remained unchanged.
VHHBERT was pre-trained for 312,500 steps, which equates to 20 epochs, with a batch size of 128 on one NVIDIA Tesla V100 GPU.
The resulting VHHBERT is available on the Hugging Face Hub\footnote{COGNANO/VHHBERT: \url{https://huggingface.co/COGNANO/VHHBERT}}.

\paragraph{Fine-tuning}

\begin{wrapfigure}[14]{r}{0.42\textwidth}
  \centering
  \vspace{-5mm}
  \includegraphics[width=0.42\textwidth]{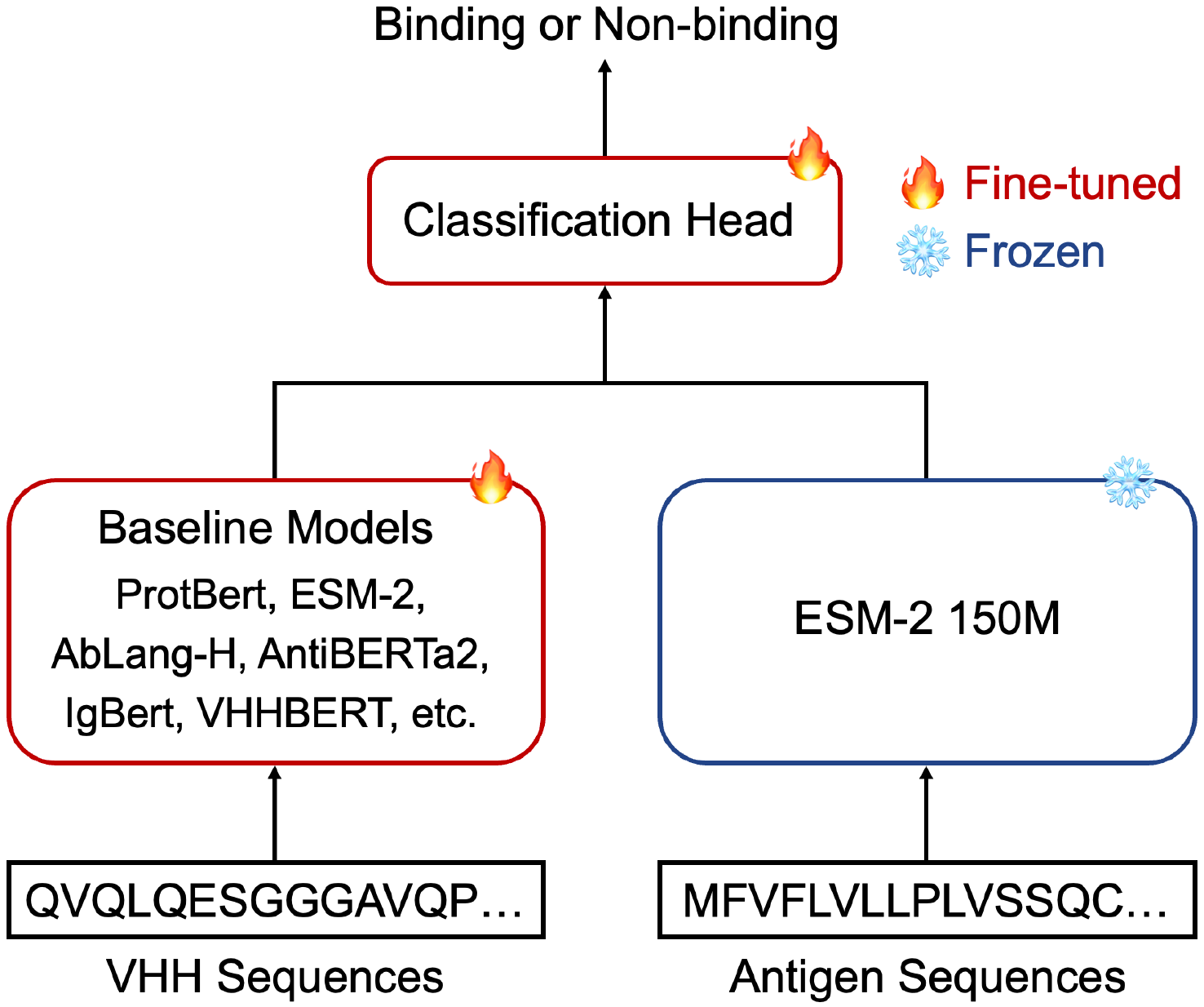}
  \caption{Overview of the experimental setup.}
  \label{fig:experimental_setup}
\end{wrapfigure}

As a fine-tuning dataset, we used AVIDa-SARS-CoV-2, which was divided by individual, as shown in Table~\ref{tab:train_test_split}.
Figure~\ref{fig:experimental_setup} shows an overview of the experimental setup.
AVIDa-SARS-CoV-2 has the amino acid sequences of VHHs and antigens as input features for binding prediction.
To obtain each sequence representation, we used the nine aforementioned baseline models for VHHs and the pre-trained protein language model ESM-2 for antigens and extracted the mean of the representations for each amino acid from the last layer in each language model.
The sequence representations of the VHHs and antigens were concatenated and utilized as input to a multi-layer perceptron, which was added on top of the two language models as a classification head.
Note that we fixed the weights of ESM-2 used for antigens and fine-tuned the classification head and the language model used for VHHs to assess the representation capabilities of antibody language models.
We trained the models for 30 epochs with a batch size of 32 on one NVIDIA Tesla V100 GPU.
We conducted five repetitive experiments with different random seeds and report the average results and standard derivation.

\subsection{Results}
\label{sec:results}

\begin{table}
  \caption{Performance comparisons of baseline models for VHH-antigen binding prediction.
  Best performance is highlighted in bold.}
  \centering
  \label{tab:benchmark_result}
  \resizebox{\textwidth}{!}{
  \begin{tabular}{lccccc}
    \toprule
    Model & Accuracy & Precision & Recall & F1-score & AUPRC \\
    \midrule
    ProtBert & 0.803~$\pm$~0.012 & 0.602~$\pm$~0.036 & 0.564~$\pm$~0.046 & 0.580~$\pm$~0.023 & 0.532~$\pm$~0.073 \\
    ESM-2 150M & 0.801~$\pm$~0.010 & 0.607~$\pm$~0.034 & 0.514~$\pm$~0.036 & 0.555~$\pm$~0.021 & 0.531~$\pm$~0.047 \\
    ESM-2 650M & 0.822~$\pm$~0.020 & 0.682~$\pm$~0.083 & 0.540~$\pm$~0.048 & 0.598~$\pm$~0.023 & 0.584~$\pm$~0.069 \\
    AbLang-H & 0.828~$\pm$~0.004 & 0.753~$\pm$~0.033 & 0.430~$\pm$~0.017 & 0.547~$\pm$~0.005 & 0.589~$\pm$~0.018 \\
    AntiBERTa2 & 0.851~$\pm$~0.007 & 0.769~$\pm$~0.044 & 0.551~$\pm$~0.021 & 0.641~$\pm$~0.008 & 0.660~$\pm$~0.018 \\
    AntiBERTa2-CSSP & \textbf{0.854~$\pm$~0.007} & 0.773~$\pm$~0.030 & 0.565~$\pm$~0.014 & \textbf{0.652~$\pm$~0.014} & \textbf{0.690~$\pm$~0.011} \\
    IgBert & 0.845~$\pm$~0.007 & 0.741~$\pm$~0.045 & 0.558~$\pm$~0.045 & 0.634~$\pm$~0.018 & 0.610~$\pm$~0.044 \\
    VHHBERT & 0.823~$\pm$~0.011 & 0.658~$\pm$~0.042 & \textbf{0.567~$\pm$~0.025} & 0.608~$\pm$~0.012 & 0.650~$\pm$~0.025 \\
    VHHBERT w/o PT & 0.831~$\pm$~0.003 & \textbf{0.811~$\pm$~0.024} & 0.392~$\pm$~0.010 & 0.528~$\pm$~0.008 & 0.624~$\pm$~0.008 \\
    \bottomrule
  \end{tabular}
  }
\end{table}

Table~\ref{tab:benchmark_result} shows the performance comparisons of the baseline models for the VHH-antigen binding prediction.
We used precision, recall, F1-score, and area under the precision-recall curve (AUPRC) in addition to accuracy as evaluation metrics because the prediction of antibody binders, which are fewer in number than non-binders, is much more important for drug discovery.
VHHBERT w/o PT showed high precision but significantly lower recall than the other models, resulting in the lowest F1-score.
This result indicates the effectiveness of pre-training on protein and antibody sequences in predicting VHH binding.
AntiBERTa2, AntiBERTa2-CSSP, IgBert, and VHHBERT pre-trained on antibody sequences outperformed ESM-2 and ProtBert pre-trained on protein sequences in accuracy, F1-score, and AUPRC.
This is consistent with previous studies~\cite{leem2022deciphering,wang2022pre,barton2024enhancing} that reported that using antibodies for pre-training, rather than general proteins, contributes to the performance of antibody-specific tasks.
In general, the antigen to which an antibody binds is determined by the amino acid sequence in CDRs.
Because CDRs are highly variable owing to mechanisms such as immunoglobulin gene rearrangement and somatic hypermutation~\cite{chi2020v}, they do not follow the evolutionary information stored in general protein sequences~\cite{vishwakarma2022vhh}.
Because of these differences in evolutionary processes, pre-training with antibody sequences should be effective in predicting VHH binding.

AntiBERTa2, AntiBERTa2-CSSP, and IgBert outperformed the other models in terms of accuracy and F1-score, suggesting that pre-training with a larger number of antibody sequences contributes to the generalization to unknown antibody clusters.
Interestingly, additional pre-training of AntiBERTa2-CSSP using human antibody structures contributed to improved performance in predicting VHH-antigen binding.
However, the highest F1-score of AntiBERTa2-CSSP remains at approximately 65\%; therefore, there is still room for performance improvement for practical drug discovery applications.
Although VHHBERT was pre-trained with significantly fewer antibody sequences than the other pre-trained antibody language models, its F1-score was higher than AbLang-H and close to IgBert.
This result can probably be attributed to differences in the sequence patterns between conventional antibodies and VHHs.
Specifically, the average length of CDR3, which is the most important for antigen recognition, is approximately 1.5 times longer for VHHs than for conventional antibodies~\cite{vu1997comparison,henry2018antigen}.
Moreover, the genetic sequences of antibodies differ between species, resulting in differences in amino acid sequences, even in non-variable regions~\cite{de2015structural,sinkora2022comparative}.
In conclusion, these insights obtained through benchmarks underscore the significance of AVIDa-SARS-CoV-2 as a useful benchmark for assessing the representation capabilities of antibody language models for binding prediction, thereby promoting the advancement of AI-assisted antibody discovery.

\section{Discussion}

\subsection{VHH-specific Language Models}

Recent remarkable progress in language models has led to the active development of domain-specific language models in various application fields~\cite{beltagy2019scibert,gu2021domain,luo2022biogpt}.
Here, we discuss the significance of building VHH-specific language models from the perspective of drug discovery.
VHHs have recently attracted attention as therapeutic agents because of their small size, high stability, good human tolerability, and relative ease of production~\cite{jovvcevska2020therapeutic,jin2023nanobodies}.
Furthermore, VHHs possess favorable properties for the construction of large-scale language models.
First, VHHs have a simple structure consisting of only heavy chains, which allows for easier identification of full-length amino acid sequences using DNA sequencing technologies.
Second, VHH acts as a single functional unit, meaning that VHH sequences contain all the information necessary for the function of an antibody against an antigen.
In contrast, conventional antibodies are composed of two pairs of heavy and light chains, and they function as a single functional unit by combining heavy and light chains.
Therefore, paired sequences should ideally be used as the input data for language models.
However, it is difficult to construct a large-scale database of paired sequences because obtaining them requires time-consuming experiments.
Indeed, the number of paired sequences recorded in the OAS database is approximately one thousand times less than that of unpaired sequences.
Accordingly, the existing antibody language models are trained primarily on unpaired heavy- and/or light-chain sequences, ignoring the effects of their counterpart chains.
Although AntiBERTa2 and IgBert used paired sequences in their training data, the majority of the training data consists of unpaired sequences.
The advantages of VHH over conventional antibodies will facilitate the construction of large-scale databases that are meaningful for therapeutic antibody discovery and pave the way for future construction of practical VHH-specific language models.

\subsection{Negative Societal Impacts}
\label{sec:negative_impacts}

The VHH binders in AVIDa-SARS-CoV-2 have the potential to be useful in COVID-19 therapeutics.
In addition, using our dataset to develop predictive models for binding to SARS-CoV-2 variants may accelerate the development of therapeutics against emerging variants of concern (VOCs).
To reap these benefits, there is a possibility that third-party organizations could use our dataset for commercial purposes.
Because this creates the risk of future conflicts of interest between third-party organizations, we have prohibited commercial use by licensing the dataset.
Furthermore, antibodies usually act as inhibitors or neutralizers of their target antigen, but some, although relatively rare, can stimulate or enhance the function of the target~\cite{schardt2022agonist,leitner2023fcgammar}.
Even if antibodies that bind to spike proteins can be identified or predicted, the possibility of promoting infection cannot be excluded.
Therefore, validation experiments and clinical trials must be conducted to confirm the usefulness of each VHH.

\subsection{Limitations and Future Work}
\label{sec:limitations}

Our dataset potentially contains data biases derived from the specific alpacas used for dataset generation.
As shown in Figure~\ref{fig:avida_dataset_analysis}(b), each individual produced biased SARS-CoV-2-specific VHHs.
This limitation may reduce the generalization performance of the models trained on our dataset, potentially hindering their practical application in antibody discovery.
The best way to address this limitation is to generate datasets from multiple individuals with different VHH gene sequences at multiple times of immunization.
This would mitigate data biases in the dataset and help a model to understand the universal knowledge of antigen-antibody interactions.
However, realistically, there are limitations in creating datasets under various conditions, owing to cost and time constraints.
Therefore, it is also necessary to develop models that can overcome data biases.
Our benchmark task serves as a valuable benchmark for evaluating whether models can overcome individual-induced biases, and to our knowledge, no other such datasets exist.
In the future, in addition to generating and publishing more diverse datasets, we plan to research further model architectures and pre-training methods to achieve a high generalization performance in real-world antibody discovery tasks.

\section{Conclusion}

In this study, we introduced AVIDa-SARS-CoV-2, a labeled dataset of SARS-CoV-2-VHH interactions, and VHHCorpus-2M, which contains over two million VHH sequences, providing novel datasets for the evaluation and pre-training of antibody language models.
In addition, we developed a VHH-specific language model, VHHBERT, pre-trained on VHHCorpus-2M and reported benchmark results for binding prediction using existing general protein and antibody-specific language models.
We envision that the availability of AVIDa-SARS-CoV-2 and VHHCorpus-2M will facilitate further research on antibody language models and their application in therapeutic antibody discovery.

\newpage

\begin{ack}

This study was supported by funding from GREEN CORE, LTD. (Tokyo, Japan), Yamato scientific group holdings Inc. (Tokyo, Japan), and POPURI Pharmacy Co., Ltd. (Kyoto, Japan) and subsidies from KYOTO Industrial Support Organization 21 (Sangakukou no Mori).
This study was also supported by the Ministry of Economy, Trade and Industry (METI) R\&D Support Program for Growth-oriented Technology SMEs Grant Number JPJ005698.
  
We thank the Japan External Trade Organization (JETRO) for supporting COGNANO Inc.
We also thank Tomohisa Oda from COGNANO Inc. and SAKURA internet Inc. for developing the websites for the publication of AVIDa-SARS-CoV-2 and VHHCorpus-2M.

\end{ack}

\bibliographystyle{splncs04}

\newpage

\appendix

\section{Appendix}

\subsection{Ethics Statement for Animal Experiments}
\label{sec:appendix_ethics}

All animal experiments on the alpacas were conducted in accordance with the KYODOKEN Institute for Animal Science Research and Development (Kyoto, Japan) and the ARRIVE (Animal Research: Reporting of \textit{In Vivo} Experiments) guidelines\footnote{ARRIVE guidelines: \url{https://arriveguidelines.org}}.
Veterinarians performed breeding, health maintenance, and immunization by adhering to the published Guidelines for Proper Conduct of Animal Experiments by the Science Council of Japan.
The KYODOKEN Institutional Animal Care and Use Committee approved the protocols for these studies (approval number 20200312).

Our data generation method uses animal models immunized with a target protein that could potentially harm the animals, such as a particular toxin, pathogen, or allergen.
Hence, the risk to the animal should be minimized by treating the immune source to inactivate or detoxify it.

\subsection{Dataset Generation}
\label{sec:appendix_dataset_generation}

Here, we describe the detailed experimental procedures and conditions.

\paragraph{Immunization}
We immunized two alpacas, Alpaca P and Alpaca C, with various purified recombinant SARS-CoV-2 spike trimers: WT-cryo, the ectodomain of the wild-type (WT) SARS-CoV-2 spike protein analyzed by cryo-EM~\cite{walls2020structure}, whose S1/S2 region is altered to avoid cleavage by furin; WT-ecto, the restored version of the furin site of the WT-cryo; WT, the full-length of the SARS-CoV-2 spike protein (GenBank: QHD43416); and S2-domain, the S2 domain of the WT.
The antigen cocktail mixture, emulsified in complete Freund's adjuvant, was injected subcutaneously into the two alpacas nine times at two-week intervals.
Lymph node and blood samples were collected multiple times, resulting in a total of 27 libraries.
After this period, we continued immunization with various SARS-CoV-2 spike proteins harboring representative mutations for the Alpha, Beta, Gamma, Delta, Kappa, Lambda, and Omicron variants.
As a result, we performed eight additional immunizations with these variants and generated four additional libraries from the final harvest samples.

\paragraph{Phage Library Construction}
Peripheral blood mononuclear cells (PBMCs) were obtained from blood samples by sucrose density gradient centrifugation using Ficoll (Nacalai Tesque, Kyoto, Japan).
The lymph nodes and PBMC samples were washed with phosphate-buffered saline (PBS, Nacalai Tesque) and suspended in an RNAlater solution (Thermo Fisher Scientific K.K., Tokyo, Japan).
Total RNA was isolated from these samples by using Direct-Zol RNA MiniPrep (Zymo Research, Irvine, CA).
Complementary DNA was synthesized from 1 $\mu$g of total RNA as a template by using random hexamer primers and SuperScript II reverse transcriptase (Thermo Fisher Scientific K.K.).
The coding regions of the VHH domain were amplified using LA Taq polymerase (TAKARA Bio Inc., Shiga, Japan) with two PAGE-purified primers (CALL001, 5'-GTCCTGGCTGCTCTTCTACAAGG-3' and CALL002, 5'-GGTACGTGCTGTTGAACTGTTCC-3'), and they were separated on a 1.5\% low-melting-temperature agarose gel (Lonza Group AG, Basel, Switzerland).
Approximately 700 base-pair bands were extracted using a QIAquick Gel Extraction Kit (Qiagen K.K., Tokyo, Japan).
Nested PCR was performed to amplify the VHH genes by using two primers that contained flanking PstI (forward) and BstEII (reverse) restriction sites to enable cloning into the pMES4 phagemid vector with a C-terminal His-tag.
Electroporation-competent Escherichia coli TG1 cells (Agilent Technologies Japan, Ltd., Tokyo, Japan) were transformed with the ligated plasmids under chilled conditions (Bio-Rad Laboratories, Inc., Hercules, CA).
The library densities were monitored and maintained at >$10^7$ colony-forming units per microliter with limiting dilution.
Colonies from 8 mL of cultured cells were harvested, pooled, and reserved in frozen glycerol stock as a mother library.
Thus, the 31 phagemid libraries were designated as the mother libraries.

\paragraph{Affinity Selection}

\begin{table}
  \caption{Details of antigen types used for dataset generation.
  Amino acid sequence for each antigen type is available at \url{https://huggingface.co/datasets/COGNANO/AVIDa-SARS-CoV-2}.
  }
  \centering
  \renewcommand{\arraystretch}{1.2}
  \label{tab:details_antigen_list}
  \resizebox{\textwidth}{!}{
  \begin{tabular}{ccll}
    \toprule
    Antigen Type & Panning & ~Mutation & ~Description \\
    \midrule
    WT & cell &
    \begin{tabular}{l}
      -
    \end{tabular} &
    \begin{tabular}{l}
      Wild-type (WT) SARS-CoV-2 identified in Wuhan with a C9 tag at the C-terminus.
    \end{tabular} \\ \hline
    D614G & cell &
    \begin{tabular}{l}
      D614G
    \end{tabular} & 
    \begin{tabular}{l}
      Mutant with D614G mutation with a C9 tag at the C-terminus.
    \end{tabular} \\ \hline
    Alpha & cell &
    \begin{tabular}{l}
      N501Y, D614G
    \end{tabular} &
    \begin{tabular}{l}
      Mutant with representative mutations of Alpha variant with a C9 tag at the C-terminus.
    \end{tabular} \\ \hline
    Alpha & bead &
    \begin{tabular}{l}
      N501Y, D614G, K986P, V987P
    \end{tabular} &
    \begin{tabular}{l}
      Mutant with representative mutations of Alpha variant with a 6$\times$His tag at the C-terminus.\\
      The PP mutation stabilizes the trimer structure~\cite{pallesen2017immunogenicity}.\\
      The sequence after the transmembrane domain is replaced by the foldon trimerization motif~\cite{miroshnikov1998engineering}.
    \end{tabular} \\ \hline
    Alpha+K417N & cell &
    \begin{tabular}{l}
      K417N, N501Y, D614G
    \end{tabular} &
    \begin{tabular}{l}
      Mutant of antigen type ``Alpha'' with K417N mutation with a C9 tag at the C-terminus.
    \end{tabular} \\ \hline
    Alpha+E484K & cell &
    \begin{tabular}{l}
      E484K, N501Y, D614G
    \end{tabular} &
    \begin{tabular}{l}
      Mutant of antigen type ``Alpha'' with E484K mutation with a C9 tag at the C-terminus.
    \end{tabular} \\ \hline
    Beta & cell &
    \begin{tabular}{l}
      K417N, E484K, N501Y, D614G
    \end{tabular} &
    \begin{tabular}{l}
      Mutant with representative mutations of Beta variant with a C9 tag at the C-terminus.
    \end{tabular} \\ \hline
    Beta & bead &
    \begin{tabular}{l}
      K417N, E484K, N501Y, D614G, K986P, V987P
    \end{tabular} &
    \begin{tabular}{l}
      Mutant with representative mutations of Beta variant with a 6$\times$His tag at the C-terminus.\\
      The PP mutation stabilizes the trimer structure~\cite{pallesen2017immunogenicity}.\\
      The sequence after the transmembrane domain is replaced by the foldon trimerization motif~\cite{miroshnikov1998engineering}.
    \end{tabular} \\ \hline
    Delta & cell &
    \begin{tabular}{l}
      L452R, T478K, D614G
    \end{tabular} &
    \begin{tabular}{l}
      Mutant with representative mutations of Delta variant with a C9 tag at the C-terminus.
    \end{tabular} \\ \hline
    Delta & bead &
    \begin{tabular}{l}
      L452R, T478K, D614G
    \end{tabular} &
    \begin{tabular}{l}
      Mutant with representative mutations of Delta variant with a 6$\times$His tag at the C-terminus.\\
      The PP mutation stabilizes the trimer structure~\cite{pallesen2017immunogenicity}.\\
      The sequence after the transmembrane domain is replaced by the foldon trimerization motif~\cite{miroshnikov1998engineering}.
    \end{tabular} \\ \hline
    Kappa & bead &
    \begin{tabular}{l}
      L452R, E484Q, D614G, K986P, V987P
    \end{tabular} &
    \begin{tabular}{l}
      Mutant with representative mutations of Kappa variant with a 6$\times$His tag at the C-terminus.\\
      The PP mutation stabilizes the trimer structure~\cite{pallesen2017immunogenicity}.\\
      The sequence after the transmembrane domain is replaced by the foldon trimerization motif~\cite{miroshnikov1998engineering}.
    \end{tabular} \\ \hline
    Lambda & bead &
    \begin{tabular}{l}
      G75V, T76I, S247\_D253del, L452Q,\\F490S, K986P, V987P
    \end{tabular} &
    \begin{tabular}{l}
      Mutant with representative mutations of Lambda variant with a 6$\times$His tag at the C-terminus.\\
      The PP mutation stabilizes the trimer structure~\cite{pallesen2017immunogenicity}.\\
      The sequence after the transmembrane domain is replaced by the foldon trimerization motif~\cite{miroshnikov1998engineering}.
    \end{tabular} \\ \hline
    Omicron & cell &
    \begin{tabular}{l}
      A67V, H69del, V70del, T95I, G142D,\\V143\_Y145del, N211I, L212V, V213P, R214E,\\G339D, S371L, S373P, S735F, K417N,\\N440K, G446S, S477N, T478K, E484A,\\Q493R, G496S, Q498R, N501Y, Y505H,\\T547K, D614G, H655Y, N679K, P681H,\\N764K, D796Y, N856K, Q954H, N969K, L981F
    \end{tabular} &
    \begin{tabular}{l}
      Mutant with representative mutations of Omicron (BA.1) variant with a C9 tag at the C-terminus.
    \end{tabular} \\ \hline
    Omicron & bead &
    \begin{tabular}{l} 
      A67V, H69del, V70del, T95I, G142D,\\V143\_Y145del, N211I, L212V, V213P, R214E,\\G339D, S371L, S373P, S735F, K417N,\\N440K, G446S, S477N, T478K, E484A,\\Q493R, G496S, Q498R, N501Y, Y505H,\\T547K, D614G, H655Y, N679K, P681H,\\N764K, D796Y, K986P, V987P
    \end{tabular} &
    \begin{tabular}{l}
      Mutant with representative mutations of Omicron (BA.1) variant with a 6$\times$His tag at the C-terminus.\\
      The PP mutation stabilizes the trimer structure~\cite{pallesen2017immunogenicity}.\\
      The sequence after the transmembrane domain is replaced by the foldon trimerization motif~\cite{miroshnikov1998engineering}.
    \end{tabular} \\ \hline
    PMS & bead &
    \begin{tabular}{l}
      L18F, V47M, H69del, V70del, D80A,\\Y145del, D215G, L242\_L244del, W258R, R346S,\\K417N, N440K, V445E, L455R, A475V,\\E484K, N501Y, D614G, A701V, P792H,\\N801D, K986P, V987P 
    \end{tabular} &
    \begin{tabular}{l}
      Polymutant spike (PMS) protein~\cite{schmidt2021high} with a 6$\times$His tag at the C-terminus.\\
      The PP mutation stabilizes the trimer structure~\cite{pallesen2017immunogenicity}.\\
      The sequence after the transmembrane domain is replaced by the foldon trimerization motif~\cite{miroshnikov1998engineering}.
    \end{tabular} \\ \hline
    S2-domain & bead &
    \begin{tabular}{l}
      K986P, V987P
    \end{tabular} &
    \begin{tabular}{l}
      The S2 domain of the WT with a 6$\times$His tag at the C-terminus.\\
      The first thirty amino acids are replaced by the IL-2 secretion signal peptide~\cite{zhang2005alteration} and a linker sequence.\\
      The PP mutation stabilizes the trimer structure~\cite{pallesen2017immunogenicity}.\\
      The sequence after the transmembrane domain is replaced by the foldon trimerization motif~\cite{miroshnikov1998engineering}.
    \end{tabular} \\ \hline
    OC43 & bead &
    \begin{tabular}{l}
      -
    \end{tabular} &
    \begin{tabular}{l}
      Human coronavirus OC43 (HCoV-OC43) with a 6$\times$His tag at the C-terminus.
    \end{tabular} \\
    \bottomrule
  \end{tabular}
  }
\end{table}

We adopted two methods for affinity selection: the use of purified ectodomains coated on beads (the bead panning method) and the use of cultured cells overexpressing full-length spike proteins (the cell panning method).
This is because we think that the bead panning method tends to give clearer signals than the cell panning method, while the latter reflects a more native three-dimensional structure as multimers.
As shown in Table~\ref{tab:details_antigen_list}, the antigens used in this procedure contain various combinations of mutations at clinically important amino acids that belong to the receptor binding domain (RBD) such as K417, L452, T478, E484 and N501.
For the bead panning method, K986P and V987P substitutions as well as the foldon trimerization motif~\cite{miroshnikov1998engineering} at the C-terminus were introduced into the antigens.
To distinguish nonspecific signals, (i) mock, (ii) APOBEC3G, (iii) homemade IgM were used as controls.
Target antigens were coupled to N-hydroxysuccinimide (NHS)-activated magnet beads (Dynabeads, Thermo).
One round of biopanning was performed using each target protein-coated magnet beads in 50 mM of phosphate buffer (pH 7.4) containing 1\% n-dodecyl-$\beta$-D-maltopyranoside (DDM: Nacalai), 0.1\% 3-[(3-cho-lamidopropyl)dimethylammonio]-1-propane sulfonate (CHAPS: Nacalai), 0.001\% cholesterol hydrogen succinate (CHS: Tokyo Chemical Industry Co., Ltd. (TCI), Tokyo, Japan), 0.1\% LMNG (Anatrace, Maumee, OH), and 500 mM of NaCl.
After three washes with the same buffer, the remaining phages bound to the beads were eluted with a trypsin-ethylenediaminetetraacetic acid (EDTA, Nacalai Tesque) solution at room temperature for 30 minutes.
For the cell panning method, one round of biopanning was performed using PFA-fixed cultured cells.
Phages dissolved in PBS containing 0.05\% Tween 20, 0.5 mM of NaCl, and 0.3\% bovine serum albumin (BSA, Nacalai) were incubated with PFA-fixed cells expressing target spike proteins for 30 minutes at room temperature.
The cells were washed three times with PBS containing 0.05\% Tween 20, 0.5 mM of NaCl, and 0.3\% BSA; the remaining phages bound to the cells were eluted with a trypsin-ethylenediaminetetraacetic acid (EDTA) solution (Nacalai) for 30 minutes at room temperature with gentle agitation.
For both panning methods, the eluate was neutralized with a PBS-diluted protein inhibitor cocktail (cOmplete, EDTA-free, protease inhibitor cocktail tablets, Roche Diagnostics GmbH, Mannheim, Germany) and used to infect electroporation-competent cells.
The infected cells were cultured in LB Miller broth containing 100 $\mu$g/mL of ampicillin (Nacalai Tesque) at 37 ${}^\circ$C overnight.
The genes of the phagemids selected by biopanning were collected with a QIAprep Miniprep Kit (Qiagen), amplified by PCR, and purified using AMPure XP beads (Beckman Coulter, High Wycombe, UK).
Then, dual-indexed libraries were prepared and sequenced on an Illumina MiSeq (Illumina, San Diego, CA) by using a MiSeq Reagent Kit v3 with paired-end 300-bp reads (Bioengineering Lab. Co., Ltd., Kanagawa, Japan).

\paragraph{Sequence Analysis}
Approximately 100,000 paired reads for each library were generated by NGS analysis.
The raw read data were trimmed to remove the adaptor sequence by using Cutadapt v3.5~\cite{martin2011cutadapt} and to remove low-quality reads by using Trimmomatic v0.39~\cite{bolger2014trimmomatic}.
The remaining paired reads were merged using fastq-join~\cite{aronesty2013comparison}, and then the VHH coding sequences were extracted using SeqKit v2.2.0~\cite{shen2016seqkit}.
The DNA sequences were translated to amino acid sequences with EMBOSS v6.6.0~\cite{rice2000emboss}, and the VHH sequences were cropped from start to stop codon.
Finally, each phagemid library was converted to a FASTA file containing tens of thousands of VHH sequences.

\paragraph{Data Labeling}

Data labeling was performed using our proposed method~\cite{tsuruta2023avida}, whose reliability was fully verified using immunofluorescence staining analysis and biolayer interferometry analysis.
The code for the data labeling is available at \url{https://github.com/cognano/AVIDa-SARS-CoV-2}.
Our labeling method determines whether a VHH binds to each antigen by applying a statistical test for differences in the proportions of each VHH in a library before and after panning.
Let $p_1$ and $p_2$ denote the population proportions of a specific VHH in the libraries before and after panning.
Moreover, let $n_1$ and $n_2$ denote the libraries' total read counts before and after panning, respectively, and let $x_1$ and $x_2$ denote the read counts of a specific VHH in the libraries.
Then, the respective sample proportions of a specific VHH in each library are $\hat{p}_1=\frac{x_1}{n_1}$ and $\hat{p}_2=\frac{x_2}{n_2}$.
Given that the minimum value of all possible $n_1$ and $n_2$ was over 5,000, we assumed that $\hat{p}_1$ and $\hat{p}_2$ follow normal distributions with mean $p_1$ and $p_2$ and variance $\frac{p_1(1-p_1)}{n_1}$ and $\frac{p_2(1-p_2)}{n_2}$, respectively, according to the central limit theorem.
Furthermore, the difference in the proportions $\hat{p}_1-\hat{p}_2$ can also be approximated by a normal distribution due to the reproductive property of the normal distribution.
Thus, the test statistic $Z$ under null hypothesis $H_0: p_{1}=p_{2}$ was calculated as follows.

\begin{equation}
  \label{eq:z_score}
  Z=\frac{\hat{p}_1-\hat{p}_2}{\sqrt{p(1-p)(\frac{1}{n_1}+\frac{1}{n_2})}}
\end{equation}

where $p$ is the pooled proportion calculated as $p=\frac{x_1+x_2}{n_1+n_2}$.
The p-value of $Z$ was calculated using the standard normal distribution.
In the same way, p-values were calculated for all VHH-target pairs in the sublibraries with respect to the corresponding mother libraries.
We adopted the smallest p-value, indicating the most significant difference in proportion, among identical VHH-target pairs.
If a specific VHH's proportion in a sublibrary increased from the proportion in the corresponding mother library and the p-value was 0.05 or less (our chosen significance level), the VHH-target pair was labeled with ``binder.''
Similarly, if the proportion decreased and the p-value was 0.05 or less, the pair was labeled with ``non-binder.''
Finally, if the p-value exceeded 0.05, the pair was labeled with ``non-significant.''
This label was excluded from AVIDa-SARS-CoV-2.

The results of biological experiments always contain background noise, such as that due to binding to contaminating proteins.
Therefore, we applied a novel noise reduction algorithm to avoid false positives and improve label reliability.
We reconfirmed VHHs labeled as a ``binder'' to any of the antigen types by comparing the labels to negative control samples under the following conditions.

\begin{enumerate}
  \item If the VHH was a non-binder to a negative control sample, the label remained ``binder.''
  \item If the VHH was a binder to a negative control sample, the label was reassigned from ``binder'' to ``noise'' because of possible false positives.
  This label was excluded from AVIDa-SARS-CoV-2.
  \item If the VHH was ``non-significant'' with respect to a negative control sample, the ratio of the p-value of the negative control sample to that of the antigen was compared to $10^{2.5}$.
  This value was empirically determined by an author (a biologist) according to feedback from biological experiments in our previous studies~\cite{maeda2022panel}.
  \begin{enumerate}
    \item If the ratio of p-values was below $10^{2.5}$, the label was reassigned from ``binder'' to ``non-significant'' because of possible false positives.
    \item If the ratio of p-values was $10^{2.5}$ or more, the label remained ``binder.''
  \end{enumerate}
\end{enumerate}

\subsection{Dataset Analysis}
\label{sec:appendix_dataset_analysis}

\begin{figure}
  \centering
  \includegraphics[width=\textwidth]{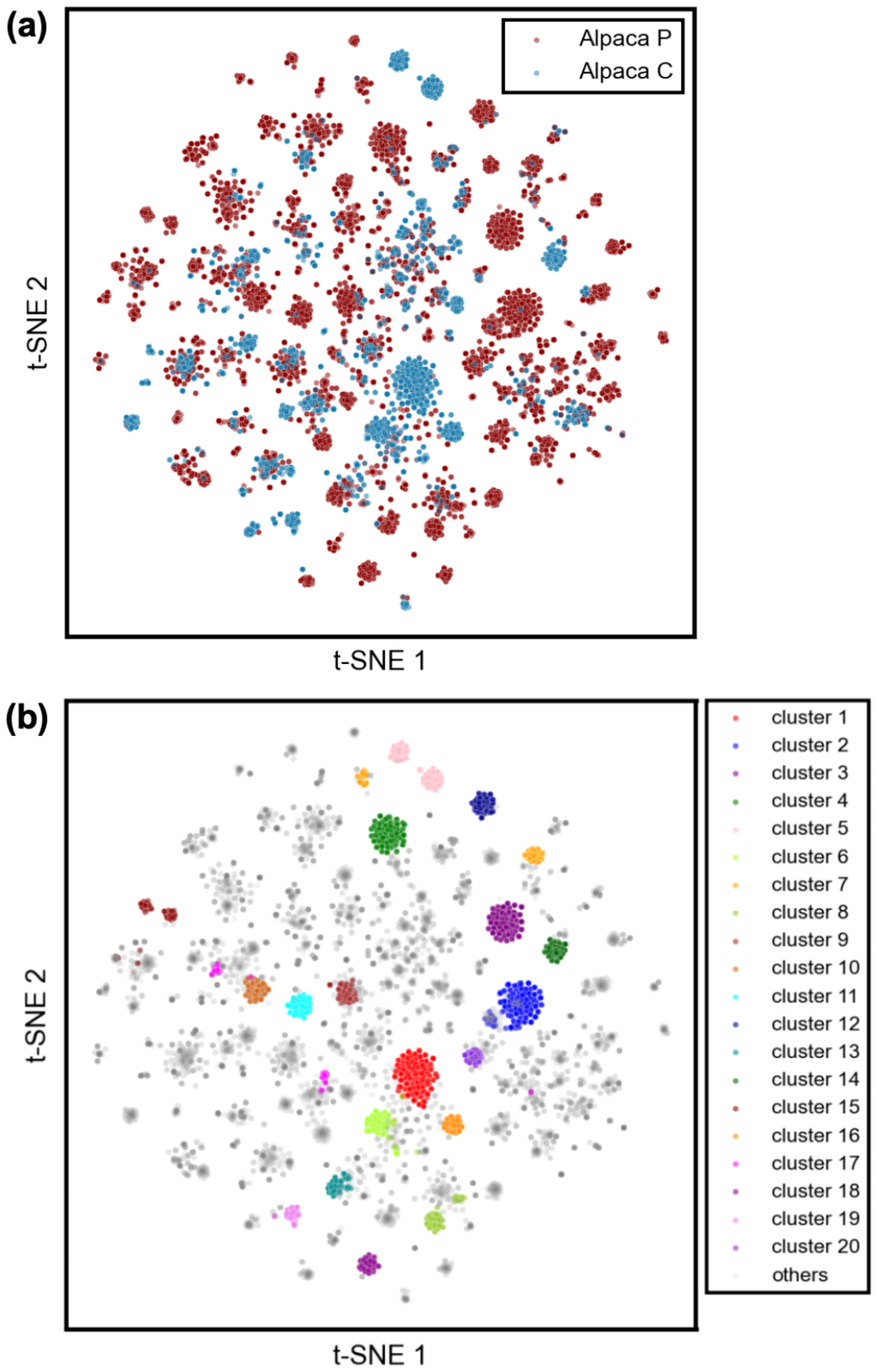}
  \caption{Two-dimensional representation of binder sequences colored by (a) individuals and (b) clusters.}
  \label{fig:vhh_2d_maps}
\end{figure}

Figure~\ref{fig:vhh_2d_maps} is enlarged version of Figure~\ref{fig:avida_dataset_analysis}(b) and (c).

\paragraph{Sequence Clustering}

In Section~\ref{sec:avida_dataset_analysis} and \ref{sec:corpus_dataset_analysis}, we performed clustering of VHH sequences using MMseqs2 v15-6f452~\cite{steinegger2017mmseqs2}\footnote{MMseqs2: \url{https://github.com/soedinglab/MMseqs2}}, released under the GPL-3.0 license.
We used the default parameters of easy-cluster, as shown in the following command.

\begin{verbatim}
mmseqs easy-cluster input.fasta clusterRes tmp --min-seq-id 0.9
\end{verbatim}

All input and output files are available at the GitHub repository\footnote{AVIDa-SARS-CoV-2: \url{https://github.com/cognano/AVIDa-SARS-CoV-2}}.

\paragraph{Dimensionality Reduction}

To obtain a 2D representation of the VHH binders as shown in Figure~\ref{fig:avida_dataset_analysis}(b) and (c), we encoded VHH sequences using Kidera factors~\cite{kidera1985statistical}, which represent the physicochemical properties of amino acids in a 10-dimensional vector, and then converted them into 2D vectors using t-SNE~\cite{van2008visualizing}.
To run t-SNE, we used sklearn.manifold.TSNE from scikit-learn v1.5.0\footnote{scikit-learn: \url{https://github.com/scikit-learn/scikit-learn}}, released under the BSD 3-Clause License, with default parameters, e.g., perplexity=30 and n\_iter=1000.

\paragraph{Pairwise Sequence Identities}

To calculate pairwise sequence identities within AVIDa-SARS-CoV-2 and VHHCorpus-2M, as shown in Figure~\ref{fig:corpus_dataset_analysis}, we used the pairwise2.align.globalxx function of Biopython v1.83~\cite{cock2009biopython}\footnote{Biopython: \url{https://github.com/biopython/biopython}}, released under the BSD 3-Clause License.

\subsection{Benchmarks}
\label{sec:appendix_benchmarks}

The code to run the pre-training of VHHBERT and fine-tuning of all baseline models is available at \url{https://github.com/cognano/AVIDa-SARS-CoV-2}.

\subsubsection{Model Implementations}

\begin{table}
  \caption{Comparison of model parameters of baseline models.
  ``M'' stands for million.}
  \centering
  \label{tab:model_parameters}
  \resizebox{\textwidth}{!}{
  \begin{tabular}{lccccccc}
    \toprule
    Parameters & ProtBert & ESM-2 150M & ESM-2 650M & AbLang-H & 
    \begin{tabular}{c}
    AntiBERTa2\\AntiBERTa2-CSSP
    \end{tabular} & IgBert & VHHBERT \\
    \midrule
    Number of layers & 30 & 30 & 33 & 12 & 16 & 30  & 12 \\
    Number of attention heads & 16 & 20 & 20 & 12 & 16 & 16 & 12 \\
    Embedding dimension & 1024 & 640 & 1280 & 768 & 1024 & 1024 & 768 \\
    Feed-forward layer dimension & 4096 & 2560 & 5120 & 3072 & 4096 & 4096 & 3072 \\
    Number of parameters & 420M & 150M & 650M & 86M & 202M & 420M & 86M \\
    \bottomrule
  \end{tabular}
  }
\end{table}

We summarize the model parameters for all baseline models in Table~\ref{tab:model_parameters}.

\begin{itemize}
  \item \textbf{ProtBert}~\cite{elnaggar2022prottrans} is a BERT-based~\cite{devlin2018bert} model pre-trained on 216 million protein sequences in UniRef100~\cite{suzek2015uniref}.
  We used a pre-trained ProtBert\footnote{ProtBert: \url{https://huggingface.co/Rostlab/prot_bert}} released under Academic Free License v3.0 on the Hugging Face Hub.
  \item \textbf{ESM-2}~\cite{lin2023evolutionary} is a protein language model pre-trained on protein sequences in UniRef~\cite{suzek2015uniref}.
  During training, sequences are sampled with even weighting across \textasciitilde43 million UniRef50 training clusters from \textasciitilde138 million UniRef90 sequences so that over the course of training the model sees \textasciitilde65 million unique sequences.
  We used a pre-trained ESM-2 with 150\footnote{ESM-2 150M: \url{https://huggingface.co/facebook/esm2_t30_150M_UR50D}} and 650\footnote{ESM-2 650M: \url{https://huggingface.co/facebook/esm2_t33_650M_UR50D}} million parameters released under the MIT License on the Hugging Face Hub.
  \item \textbf{AbLang-H}~\cite{olsen2022ablang} is a RoBERTa-based~\cite{liu2019roberta} model pre-trained on 14 million heavy chains of antibodies in the OAS database.
  We used a pre-trained AbLang-H released in AbLang v0.3.1\footnote{AbLang: \url{https://github.com/oxpig/AbLang}}, an open source library under the BSD 3-Clause License.
  Because AbLang-H has a positional embedding layer with a maximum length of 160, we trimmed the first 10 amino acids of the input sequence, which was derived from a phagemid vector rather than a VHH, to accommodate the maximum sequence length of 166 in AVIDa-SARS-CoV-2.
  \item \textbf{AntiBERTa2}~\cite{barton2024enhancing} is a RoFormer-based~\cite{su2024roformer} model pre-trained using 823.7 million antibody sequences in the OAS and proprietary databases, consisting of 821.2 million unpaired antibody sequences and 2.5 million paired heavy and light chain antibody sequences.
  We used a pre-trained AntiBERTa2\footnote{AntiBERTa2: \url{https://huggingface.co/alchemab/antiberta2}} released under a modified version of the Apache 2.0 License\footnote{AntiBERTa2 License: \url{https://huggingface.co/alchemab/antiberta2/blob/main/LICENSE.md}} on the Hugging Face Hub.
  \item \textbf{AntiBERTa2-CSSP}~\cite{barton2024enhancing} is a multimodal version of AntiBERTa2.
  AntiBERTa2-CSSP was trained using 1,554 human antibody structures in SAbDab~\cite{dunbar2014sabdab} via contrastive sequence-structure pre-training (CSSP), which amalgamates the representations of antibody sequences and structures in a mutual latent space.
  We used a pre-trained AntiBERTa2-CSSP\footnote{AntiBERTa2-CSSP: \url{https://huggingface.co/alchemab/antiberta2-cssp}} released under a modified version of the Apache 2.0 License\footnote{AntiBERTa2-CSSP License: \url{https://huggingface.co/alchemab/antiberta2-cssp/blob/main/LICENSE.md}} on the Hugging Face Hub.
  \item \textbf{IgBert}~\cite{kenlay2024large} is a model initialized with weights of ProtBert and trained using more than two billion unpaired sequences of light and heavy chains and two million paired sequences in the OAS database.
  We used a pre-trained IgBert\footnote{IgBert: \url{https://huggingface.co/Exscientia/IgBert}} released under the MIT License on the Hugging Face Hub.
  \item \textbf{VHHBERT} is a RoBERTa-based model pre-trained on two million VHH sequences in VHHCorpus-2M.
  We implemented VHHBERT by using transformers v4.41.1\footnote{huggingface/transformers: \url{https://github.com/huggingface/transformers}} released under the Apache License 2.0.
  We used the same model parameters as RoBERTa\textsubscript{BASE}, except that it used positional embeddings with a length of 185 to cover the maximum sequence length of 179 in VHHCorpus-2M.
  We released the pre-trained VHHBERT\footnote{COGNANO/VHHBERT: \url{https://huggingface.co/COGNANO/VHHBERT}} under the MIT License on the Hugging Face Hub.
\end{itemize}

\subsubsection{Pre-training Details}

As a pre-training corpus for VHHBERT, VHHCorpus-2M was randomly divided into 2,000,000 training sets and 40,988 validation sets.
The VHH sequences were tokenized using a vocabulary file\footnote{Vocabulary file: \url{https://huggingface.co/COGNANO/VHHBERT/blob/main/vocab.txt}} consisting of 25 tokens: 20 amino acids and 5 special tokens, resulting in amino acids in the VHH sequence being mapped to different token IDs.
Each VHH sequence was padded to the maximum length in each mini-batch during training.
We used masked language modeling as the pre-training objective.
During pre-training, 15\% of the residues from each VHH sequence were randomly selected, and of these, 80\% were masked, 10\% were randomly changed to another residue, and 10\% remained unchanged.

VHHBERT was pre-trained for 312,500 steps, which equates to 20 epochs, with a batch size of 128 on a single machine with one NVIDIA Tesla V100 GPU on the SAKURA cloud\footnote{SAKURA cloud: \url{https://cloud.sakura.ad.jp}}.
The learning rate was warmed up over the first 5\% of the total steps to a peak learning rate of 1e-4 and linearly decayed thereafter.
We used the AdamW optimizer~\cite{loshchilov2017decoupled} with $\beta_{1}$~=~0.90, $\beta_{2}$~=~0.98, $\epsilon$~=~1e-6, and a weight decay of 0.01.
The training time was approximately three days.

\subsubsection{Fine-tuning Details}

As a fine-tuning dataset, we used AVIDa-SARS-CoV-2, which was divided by individual, as shown in Table~\ref{tab:train_test_split}.
AVIDa-SARS-CoV-2 has the amino acid sequences of VHHs and antigens as input features for binding prediction.
To obtain each sequence representation, we used the nine baseline models described in Section~\ref{sec:baseline_models} for VHHs and the pre-trained protein language model ESM-2 150M for antigens, respectively.
We used the mean of the representations for each amino acid from the last layer in each language model, which is independent of the sequence length, as a sequence representation.
Because the maximum length of an antigen sequence is 1328 and ESM-2 150M has a positional embedding layer with a maximum length of 1024, the antigen sequence was divided by 1000 and input into ESM-2 150M separately.
The obtained representations were concatenated and averaged to form a sequence representation of the antigen.
The sequence representations of the VHHs and antigens obtained from the language models were concatenated and utilized as input to a multi-layer perceptron with one hidden layer of 768 neurons, which was added on top of the two language models as a classification head.
Note that we fixed the weights of ESM-2 150M used for antigens and fine-tuned the classification head and the language model used for VHHs to assess the representation capabilities of antibody language models.

We trained the models for 30 epochs with a batch size of 32 on a single machine with one NVIDIA Tesla V100 GPU on the SAKURA cloud.
We performed early stopping with a patience of five epochs based on the F1-score on the validation set, which was a randomly sampled 10\% from the training set, and selected the model with the best F1-score to evaluate the model performance on the test set.
The learning rate was warmed up over the first 5\% of the total steps to a peak learning rate of 1e-6 and linearly decayed thereafter.
We used the AdamW optimizer with $\beta_{1}$~=~0.90, $\beta_{2}$~=~0.98, $\epsilon$~=~1e-6, and a weight decay of 0.01.
We conducted five repetitive experiments with different random seeds and report the average results and standard derivation.

\end{document}